\pdfoutput=1

\documentclass[11pt,xcolor=dvipsnames]{article}

\usepackage[final]{EMNLP2023}

\usepackage{times}
\usepackage{latexsym}

\usepackage[T1]{fontenc}

\usepackage[utf8]{inputenc}

\usepackage{microtype}

\usepackage{inconsolata}

\usepackage{graphicx}
\usepackage{hyperref}
\usepackage{amsmath}
\usepackage{multirow}
\usepackage{tablefootnote}
\usepackage{fontawesome5}
\usepackage{tabularx}
\usepackage[normalem]{ulem}
\usepackage{amsfonts}
\usepackage{xcolor}

\title{AdaSent: Efficient Domain-Adapted Sentence Embeddings \\ for Few-Shot Classification}

\author{Yongxin Huang$^{1}$, Kexin Wang$^{1}$, Sourav Dutta$^{2}$, \\ {\bf Raj Nath Patel$^{2}$, Goran Glavaš$^{3}$, Iryna Gurevych$^{1}$ } \\
        $^{1}$Ubiquitous Knowledge Processing Lab (UKP Lab) \\ Department of Computer Science and Hessian Center for AI (hessian.AI) \\ Technical University of Darmstadt \\ $^{2}$Huawei Research Centre, Dublin, Ireland \\ $^{3}$Center for AI and Data Science, University of Würzburg \\ $^{1}$\texttt{www.ukp.tu-darmstadt.de} \\ $^{2}$\texttt{\{sourav.dutta2,raj.nath.patel\}@huawei.com} \\ $^{3}$\texttt{goran.glavas@uni-wuerzburg.de}}

\begin{document}
\maketitle
\begin{abstract}

Recent work has found that few-shot sentence classification based on pre-trained Sentence Encoders (SEs) is efficient, robust, and effective. 
In this work, we investigate strategies for domain-specialization in the context of few-shot sentence classification with SEs. 
We first establish that unsupervised Domain-Adaptive Pre-Training (DAPT) of a base Pre-trained Language Model (PLM) (i.e., not an SE) substantially improves the accuracy of few-shot sentence classification by up to 8.4 points. 
However, applying DAPT on SEs, on the one hand, disrupts the effects of their (general-domain) Sentence Embedding Pre-Training (SEPT). On the other hand, applying general-domain SEPT on top of a domain-adapted base PLM (i.e., after DAPT) is effective but inefficient, since the computationally expensive SEPT needs to be executed on top of a DAPT-ed PLM of each domain. 
As a solution, we propose AdaSent, which decouples SEPT from DAPT by training a SEPT adapter on the base PLM. The adapter can be inserted into DAPT-ed PLMs from any domain. 
We demonstrate AdaSent's effectiveness in extensive experiments on 17 different few-shot sentence classification datasets. AdaSent matches or surpasses the performance of full SEPT on DAPT-ed PLM, while substantially reducing the training costs. The code for AdaSent is available\footnote{\url{https://github.com/UKPLab/AdaSent}}. 

\end{abstract}

\section{Introduction}

Few-shot learning aims at training an effective model with a few labeled examples, reducing the cost of developing models for new domains and tasks. 
In recent work, SetFit~\cite{tunstall2022efficient} achieves strong performance in few-shot classification by contrastively fine-tuning~\cite{koch2015siamese} pre-trained sentence embeddings. Being prompt-free and effective on relative small models, SetFit is much more efficient than popular prompt-based methods including In-Context Learning (ICL, \citealp{NEURIPS2020_1457c0d6_gpt3}) and Pattern Exploit Training (PET, \citealp{schick-schutze-2021-exploiting}), which require careful prompt engineering and large model size.  

\begin{figure}[t]
    \centering
    \includegraphics[width=0.8\columnwidth]{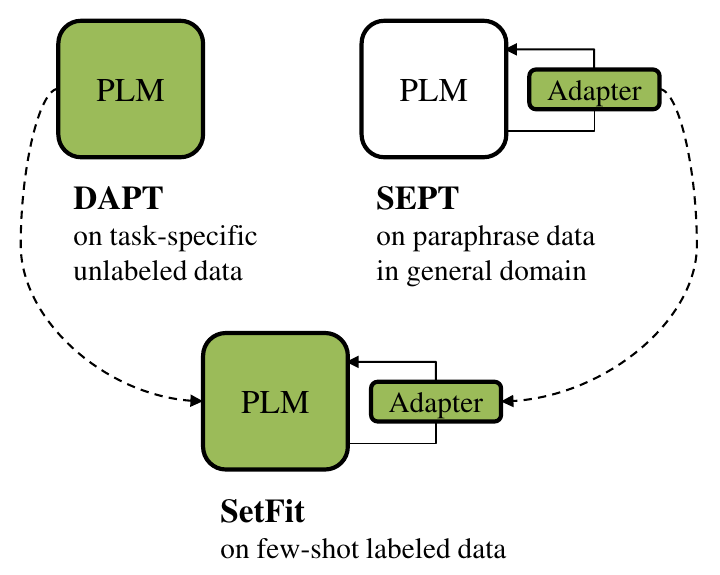}
    \caption{Training diagram of {AdaSent}. Trainable parameters are marked in \textcolor{green!40!black}{green}. After Domain-Adaptive Pre-training (DAPT) on the Pre-Trained Language Model (PLM) and Sentence-Embedding Pre-Training (SEPT) with an adapter, the two parts are assembled together to perform SetFit for few-shot classification.} 
    \label{fig:adasent}
\end{figure}

Despite its success, SetFit fine-tunes a sentence encoder with only a few labeled samples without leveraging unlabeled data from the target-task domain, which are easy to obtain. It is well-known that Domain-Adaptive Pre-Training (DAPT)\footnote{By DAPT we refer to the TAPT (Task-Adaptive Pre-Training) in~\citet{gururangan-etal-2020-dont}. We do not strictly differentiate between domain and task in the present work.}
on a vanilla PLM with unlabeled in-domain data 
can significantly improve its downstream performance~\cite{han-eisenstein-2019-unsupervised,gururangan-etal-2020-dont}. 
However, it is ineffective to apply DAPT on sentence encoders, i.e. vanilla PLMs that have undergone Sentence Embedding Pre-Training (SEPT, \citealp{reimers-gurevych-2019-sentence}) in general domain, as DAPT messes up the effects of SEPT and disrupts the model’s ability to semantically accurately embed sentences. 
Though DAPT \textit{before} SEPT is effective in contrast \citep{wang-etal-2021-tsdae-using}, it is computationally inefficient as the general-domain SEPT has to be done all over again on every domain-adapted PLM if we have more than one domain. 

To create a domain-specialized sentence encoder for few-shot sentence classification both efficiently and effectively, we propose \emph{AdaSent}, which combines DAPT and SEPT in a modular fashion. Specifically, it stores the sentence-specialization abilities – obtained via a single SEPT procedure in the general domain – into an adapter. This sentence-encoding adapter is trained once regardless of the number of domains, and can be plugged into domain-adapted PLMs from various domains to make them domain-specialized sentence encoders, on which SetFit is carried out to do downstream classification training (\autoref{fig:adasent}). 
Our experiments show that AdaSent can match or surpass the inefficient "full SEPT \textit{after} DAPT" approach's performance on 17 sentence classification tasks from various domains. The contribution of AdaSent is two-fold:
\begin{itemize}
    \item AdaSent significantly improves SetFit, the previous state-of-the-art few-shot classification approach, by leveraging unlabeled task-specific data through DAPT.                                                   
    \item AdaSent resolves the conflict between DAPT and SEPT and the efficiency issue of the sequential execution of both training procedures, by combining them in a modular fashion without sacrificing the performance. 
\end{itemize}

\section{Related Work}
\subsection{Text Classification with Sentence Embeddings}
Transformer-based~\cite{transformer} Pre-trained Language Models (PLMs) \citep{devlin-etal-2019-bert, roberta, Sanh2019DistilBERTAD} can be fine-tuned to build sentence embedding models~\citep{reimers-gurevych-2019-sentence}. Since the original goal of training sentence embeddings is to better model the sentence similarity for applications such as dense retrieval and sentence clustering \citep{reimers-gurevych-2019-sentence}, their usage is less explored in text classification. Though frozen sentence embeddings can directly serve as input features in text classification \citep{perone2018evaluation, scholarly}, the performance is limited compared to standard full fine-tuning of PLMs~\cite{lp-vs-ft}. To compensate this performance loss, \citet{meta-emb} concatenate encodings from various Sentence Transformers to form semantically richer sentence representations, achieving results comparable to standard fine-tuning, but at the cost of slower inference. More recently, SetFit \citep{tunstall2022efficient} significantly improves the few-shot classification by contrastively fine-tuning a pre-trained sentence-embedding model before training a classification head. Despite efficiently utilizing the limited labeled samples, SetFit does not leverage the abundant in-domain unlabeled data that can provide more domain knowledge for the task.

\subsection{Few-Shot Text Classification}
Large language models can perform few-shot classification through ICL with task-specific prompts consisting of a few labeled examples \citep{NEURIPS2020_1457c0d6_gpt3}. Though it avoids any gradient update, ICL relies on large model sizes for good performance, which makes inference costly. Prompt-based fine-tuning, on the other hand, can work with smaller models \citep{schick-schutze-2021-exploiting,tam-etal-2021-improving-adapet,gao-etal-2021-making_bmlff}. Parameter Efficient Fine-Tuning (PEFT) can further reduce the training cost by fine-tuning a much smaller module in a frozen PLM~\citep{pmlr-v97-houlsby19a, li-liang-2021-prefix, hu2022lora, karimi-mahabadi-etal-2022-prompt_perfect, he2022towards, NEURIPS2022_0cde695b_ia3}.
As an alternative way to employ task instructions, \citet{su2023embedder} train domain- and task-aware text embeddings by prepending instructions to the input text. In contrast to these methods, SetFit and our approach not only require a smaller model size, but also eliminate the need for prompts or instructions, which can introduce large variance and should be carefully designed~\cite{true-few-shot}.

\subsection{Domain Adaptation of Language Models}
\label{sec:da}
One typical way for creating domain-specific language models is pre-training through Masked Language Modelling on in-domain corpora, either continuously \citep{gururangan-etal-2020-dont} or from-scratch \citep{biobert}. An alternative is adapting the tokenizer to accommodate domain-specific vocabulary \citep{sachidananda-etal-2021-efficient, yao-etal-2021-adapt}.

For sentence embedding models specifically, domain adaptation is usually done through unsupervised training with novel objectives \citep{wang-etal-2021-tsdae-using, liu2022masked} or in-domain data generation \citep{wang-etal-2022-gpl}, mainly for the similarity or relevance estimation tasks. However, supervised sentence embedding training with general-domain data (SEPT) is always required \textit{after} the unsupervised domain-specific training phase (DAPT) to achieve optimal performance~\citep{wang-etal-2021-tsdae-using}. 
Our proposed method is inspired by the idea of disentangling domain adaptation and the downstream relevance estimation task via PEFT in \citet{disentangle}. In the present study, we show that PEFT can also be used to decouple DAPT and SEPT for few-shot classification tasks.

\subsection{Semi-Supervised Text Classification}
Unsupervised data can be incorporated in various ways to improve few-shot classification. While the DAPT approaches in~\autoref{sec:da} allow the model to learn domain-specific features in a task-agnostic way, other semi-supervised methods typically propagate task information from labeled data to unlabeled data through pseudo labeling. The pseudo-labeled data are either used for self-training \citep{schick-schutze-2021-exploiting} or consistency training \citep{uda}. All these approaches can also be combined to enable more efficient use of unlabeled data \citep{li-etal-2021-task-adaptive, chen-etal-2021-revisiting_sflm, zhao-yao-2022-eico}. In our experiments, we found that simple self-training using the same data for DAPT can further improve the performance of AdaSent. 

\section{Background}
\subsection{SetFit}
\label{sec:setfit}
SetFit \citep{tunstall2022efficient} is a two-step training procedure based on pre-trained sentence-embedding Transformer models for few-shot sentence classification. In the sentence-embedding fine-tuning step, positive and negative sentence pairs are generated from few-shot labeled sentences as follows: Pairs consisting of sentences from the same class are labeled positively with a score of 1 and pairs of sentences from different classes are assigned a negative score of 0. These generated pairs are used to fine-tune the sentence-embedding model with the Cosine Similarity Loss: 
$$
 L_{\mathrm{cosine}} = \left\| y - \mathrm{cos\_sim}(u, v) \right\|_2,
$$where $u,v\in \mathbb{R}^{D}$ are the $D$-dimensional sentence embeddings of two sentences respectively and $y\in\{0,1\}$ is the pair label. This aims to push instances of the same classes closer together in the representation space and those from different classes further apart, thereby clustering sentences according to their class labels to provide a clearer decision boundary for the classifier training later. In the second step, the Transformer is frozen to embed the original few-shot sentences. These sentence embeddings are used as input features to train a simple Logistic Regression~\cite{cox1958regression} classification head.

\subsection{Sentence Embedding Pre-Training (SEPT)}
\label{sec:sept}
As will be shown in~\autoref{sec:dapt_and_sept}, the success of SetFit heavily relies on SEPT. This is because the averaged word representations or the \texttt{[CLS]} representation from a PLM cannot capture the sentence semantics well without further training with sentence-level objectives~\citep{reimers-gurevych-2019-sentence}. The purpose of sentence-embedding pre-training is to train universal semantic representations that can be fine-tuned for different downstream tasks, e.g. in SetFit. Unlike SetFit, sentences with similar meaning are brought closer together in SEPT, while those with dissimilar meanings are pushed apart. Sentence pairs for this kind of contrastive training are typically obtained from Natural Language Inference (NLI, \citealp{snli,mnli}) or paraphrase datasets in the general domain. Sentence pairs labeled as "entailment" or "paraphrase" in the original datasets are used as positive pairs, i.e. sentences with similar meaning, in SEPT. The Multiple-Negative Ranking Loss (MNRL, \citealp{henderson2017efficient_mnrl}) with in-batch negatives is usually applied for training: \[ L_{\mathrm{MNRL}} = -\frac{1}{K}\sum_{i=1}^{K}\log{\frac{e^{\mathrm{cos\_sim}(x_i,y_i)}}{\sum_{j=1}^{K}e^{\mathrm{cos\_sim}(x_i, y_j)}}},\]
where $\{(x_i, y_i)\}_{i=1}^K$ are a batch of $K$ positive sentence pairs.

\begin{figure*}[t]
    \centering
    \includegraphics[width=\textwidth]{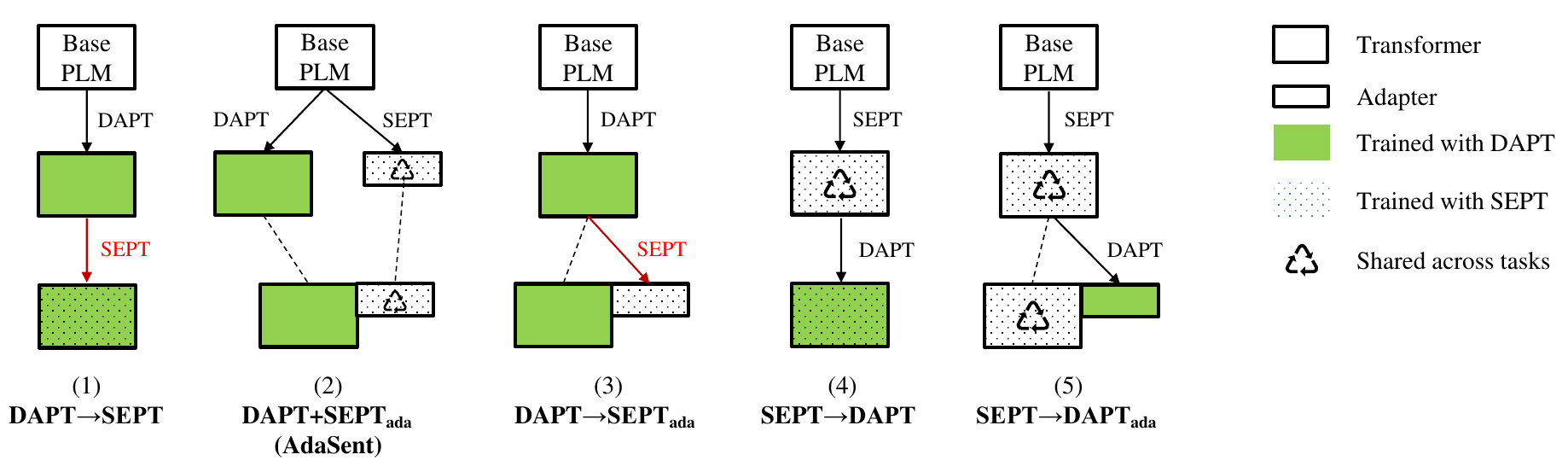}
    \caption{Five ways to combine Domain-Adaptive Pre-Training (DAPT) and Sentence Embedding Pre-Training (SEPT). An arrow pointing from a Transformer to an adapter means the adapter is trained on that Transformer. A dashed line means simple module assembly without any parameter tuning. \faRecycle{} marks trained parameters that are reusable and shared across downstream tasks. In contrast, all SEPT training starting from a DAPT Transformer (red arrows) must be repeated on every downstream task.} 
    \label{fig:models}
\end{figure*}

\subsection{Domain-Adapted Sentence Embeddings}
\label{sec:3.3}
The definition of sentence similarity varies from domain to domain, but labeled data for SEPT are usually expensive to obtain in specialized domains. \citet{wang-etal-2021-tsdae-using} found that domain-adapted sentence embedding models can be trained following a two-stage recipe: first doing unsupervised DAPT (e.g. MLM) on the domain-specific corpus, then applying supervised SEPT in the general domain (\autoref{fig:models} (1)). With this training order, if we want to train models for various domains, the same second stage has to be repeated for every domain, although it does not involve any domain-specific data. Such computational overhead cannot be avoided by simply reversing the order of the two training stages (\autoref{fig:models} (4)), since it has been shown in previous work that DAPT after the generic sentence embedding training has a negative impact on the downstream performance \citep{wang-etal-2021-tsdae-using}.

\section{Method} \label{method}

As illustrated in \autoref{fig:adasent}, our method for few-shot classification with domain-adapted sentence embeddings consists of three parts of training: (1) DAPT on the base PLM with task-specific unlabeled data, (2) SEPT on an adapter module with labeled sentence pairs from the general domain and (3) SetFit on the whole architecture (i.e. both the PLM and the adapter) with few-shot labeled data. 

In the first part, specifically, we continue to train a base PLM like DistilRoBERTa on unlabeled target task data with the MLM loss to learn domain-specific language knowledge. 
In another separate procedure, SEPT is done by tuning an adapter on a frozen base Transformer (the same PLM as in DAPT) without any domain adaptation. 
Once the domain-independent sentence encoding adapter is trained, it can be easily inserted into different DAPT models, ready for the few-shot classification task learning via SetFit in the third part.

Compared to the previous approach described in \autoref{sec:3.3}, AdaSent is more efficient for three reasons. Most significantly, our SEPT adapter is trained only once and shared across various downstream classification tasks, avoiding the overhead of repeating SEPT on new DAPT-ed models. Moreover, AdaSent allows for the independent execution of DAPT and SEPT, eliminating the need for sequential training. Therefore, they can be run concurrently in parallel to save training time. Lastly, training an adapter instead of the full model in SEPT reduces the number of trainable parameters.

\begin{table}[t]
\small
\centering
\resizebox{\columnwidth}{!}{
\begin{tabular}{lllll} 
\hline
 \textbf{Data} & \textbf{{All}}%
 & \textbf{{Paraphrase}}
 & \textbf{{NLI+SC+SE}} & \textbf{{NLI}} \\ 
\hline
\textbf{Data Size} & 1B & 86M & 0.6M & 0.3M \\
\textbf{Accuracy} & 68.6 & 70.0 & 70.0 & 68.8 \\
\hline
\end{tabular}}
\caption{SetFit accuracy on the MTEB classification tasks (see \autoref{sec:eval}) of sentence embedding models trained on different SEPT datasets without domain adaption. All and Paraphrase stand for the \emph{all-distilroberta-v1}\tablefootnote{\url{https://huggingface.co/sentence-transformers/all-distilroberta-v1}} and the \emph{paraphrase-distilroberta-base-v2}\tablefootnote{\url{https://huggingface.co/sentence-transformers/paraphrase-distilroberta-base-v2}}, respectively. }
\label{tbl:inter_table}
\end{table}

Given the extensive number of experiments in this study, we use a mixture of three datasets for SEPT, dubbed \textbf{NLI+SC+SE}, consisting SNLI \citep{snli} + MultiNLI \citep{mnli}, Sentence Compression \citep{filippova-altun-2013-overcoming_compress} and StackExchange duplicate questions, for the sake of simplicity. This is a much smaller subset of the 1 billion sentence pairs on which the popular off-the-shelf sentence embedding models\footnote{\url{https://huggingface.co/sentence-transformers}} are pre-trained. We found that these three SEPT datasets transfer the best for the downstream classification tasks\footnote{See \autoref{sec:sept_data} for results of individual SEPT datasets.}, and are adequate to train a model that performs on par with or even better than off-the-shelf models as shown in \autoref{tbl:inter_table}.

\section{Experimental Setup}

\subsection{Models}
\label{sec:models}
We experiment with three baselines and five types of domain-adapted sentence embedding models. All of these models serve as the sentence encoder in the SetFit for the few-shot classification tasks.  
The baselines are: (1) \textbf{{Base}}, the base PLM without any DAPT or SEPT; (2) \textbf{{SEPT}}, with only SEPT on the base PLM, which is also the default encoder in the original SetFit work; (3) \textbf{{DAPT}}, a domain-adapted PLM, i.e. the Base model continuously pre-trained on the in-domain corpus without SEPT. 
We also experiment with five variations of domain-adapted sentence embedding models, which differ in the way SEPT and DAPT are combined (\autoref{fig:models}). In detail, they are: 
(1) \textbf{{DAPT$\rightarrow$SEPT}}, created through DAPT followed by SEPT on the full Transformer parameters without adapter; 
(2) \textbf{{DAPT+SEPT}\textsubscript{ada}} is our AdaSent model; 
(3) \textbf{{DAPT$\rightarrow$SEPT}\textsubscript{ada}} differs from AdaSent in the training of the SEPT adapter, which is trained on the DAPT model instead of the base PLM;
(4) \textbf{{SEPT$\rightarrow$DAPT}} reverses the training order of (1), namely doing DAPT after SEPT; 
(5) \textbf{{SEPT$\rightarrow$DAPT}\textsubscript{ada}} trains a DAPT adapter on a frozen SEPT model. It requires the shortest training time, since it avoids any update of the Transformer parameters.

\subsection{Training Details}
We use {DistilRoBERTa} as the base PLM in our main experiments. Additional results on {DistilBERT} are reported in the \autoref{sec:distilbert}. We set the maximum sequence length to 512. We do not tune the hyperparameters and keep them the same for all downstream tasks. If not stated otherwise, the default setting in the used libraries (cf. \autoref{sec:code}) is applied.  
For DAPT with MLM in the main experiments, we train for a fixed number of 2344 steps\footnote{This corresponds to 3 epochs on the largest training set in our evaluation datasets.} with a batch size of 256. When using PEFT methods for DAPT, we keep the same batch size and number of steps, but with a larger learning rate of $1e-4$. 
For SEPT, we train with a batch size of 64 for 1 epoch; the learning rates are 2e-5 and 1e-4 for full and parameter-efficient training, respectively. 
For parameter-efficient training, a parallel adapter \citep{he2022towards} is used by default. We also provide results of three other different PEFT methods: bottleneck adapter \citep{pmlr-v97-houlsby19a, pfeiffer-etal-2020-mad}, LoRA \citep{hu2022lora} and prefix-tuning \citep{li-liang-2021-prefix}. 

In a separate experiment (\autoref{sec:objectives}), we compare, on models DAPT, {DAPT$\rightarrow$SEPT} and {SEPT$\rightarrow$DAPT}, three objectives for DAPT: MLM, TSDAE \citep{wang-etal-2021-tsdae-using} and SimCSE \citep{gao-etal-2021-simcse}. The latter two are designed for unsupervised sentence embedding learning, representing two mainstream training objectives for this task: denoising autoencoding and contrastive learning, respectively. For all three objectives, we train on the unlabeled dataset for 3 epochs. The batch sizes are 8, 8, 64 and the learning rates are 5e-5, 3e-5 and 1e-2, respectively. We only use NLI data in SEPT here for simplicity. The same setting is applied for the experiment in \autoref{sec:explain_reason}.

For each downstream classification task, we do SetFit on all the models with 8-shot labeled data per class for 1 epoch. The default classification head in SetFit is Logistic Regression. 

\subsection{Evaluation}
\label{sec:eval}
\begin{table*}[t]
\centering
\resizebox{\textwidth}{!}{
\begin{tabular}{llcccl} 
\hline
\textbf{Dataset} & \textbf{Abbr.} & \textbf{\# Train} & \textbf{\# Class} & \begin{tabular}[c]{@{}c@{}}\textbf{Seq. len.}\\\textbf{ (words)~}\end{tabular} & \textbf{Description} \\ 
\hline
\multicolumn{6}{c}{\textit{MTEB classification}} \\
Amazon Counterfactual & AC & 4018 & 2 & 20 & Amazon customer reviews labeled as counterfactual or not. \\
Banking77 & BANK & 10003 & 77 & 11 & Banking querys with corresponding intents. \\
Amazon Massive Intent & AMI & 11514 & 60 & 6 & Amazon Alexa utterances with associated intent. \\
Amazon Massive Scenario & AMS & 11514 & 18 & 6 & Amazon Alexa utterances with theme. \\
MTOP Intent~ & MI & 15667 & 113 & 7 & Task-oriented dialog utterances with intent. \\
MTOP Domain & MD & 15667 & 11 & 7 & Task-oriented dialog utterances with domain. \\
Emotion & EMO & 16000 & 6 & 19 & Twitter messages with basic emotion type. \\
IMDb & IMDB & 25000 & 2 & 233 & Movie reviews as positive or negative. \\
Twitter Sentiment Extraction & TSE & 27481 & 3 & 12 & Tweet sentiment classification as neutral, positive or negative. \\
Toxic Conversation & TC & 50000 & 2 & 51 & Comments from the Civil Comments platform as toxic or not. \\
Amazon Reviews Multi & ARM & 200000 & 5 & 38 & Amazon reviews with 1-5 stars. \\ 
\hline
\multicolumn{6}{c}{\textit{Domain-specific tasks}} \\
Financial PhraseBank & FPB & 3876 & 3 & 23 & Financial news
  headlines with the view of a retail investor. \\
Twitter Financial News Sentiment & TFNS & 8588 & 3 & 12 & Finance-related
  tweets with their sentiment. \\
Twitter Financial News Topic & TFNT & 15291 & 20 & 18 & Finance-related
  tweets with their topic. \\
Adverse Drug Events & ADE & 18812 & 2 & 19 & Classify
  if a sentence is ADE-related or not. \\
PubMed RCT & RCT & 176642 & 5 & 27 & PubMed abstract sentences with their role in the abstract. \\
LEDGAR & LED & 60000 & 100 & 114 & Contract provisions with their main topic. \\
\hline
\end{tabular}}
\caption{Overview of the evaluation datasets. All tasks are multi-class classification. From the training set, only 8 labeled shots per class are used for SetFit. The whole training set is used in DAPT without labels. Examples from each dataset can be found in \autoref{sec:dataset_appendix}.}
\label{tbl:dataset}
\end{table*}

We evaluate the models on 17 classification tasks, an overview of which is provided in \autoref{tbl:dataset}. These include 11 datasets from the MTEB (Massive Text Embedding Benchmark, \citealp{mteb}). For datasets that contain multilingual data, we only use the English subset in this work.  
Since most of the MTEB tasks are from the general domain, we add another six tasks for domain-specific cases, including Adverse Drug Events Binary Classification (\citealp{ade}) and PubMed RCT \citep{rct} from the biomedical domain, LEDGAR \citep{tuggener-etal-2020-ledgar,chalkidis-etal-2022-lexglue} from the legal domain, as well as Financial PhraseBank \citep{Malo2014GoodDO}, Twitter Financial News Sentiment\footnote{\url{https://huggingface.co/datasets/zeroshot/twitter-financial-news-sentiment}} and Twitter Financial News Topic\footnote{\url{https://huggingface.co/datasets/zeroshot/twitter-financial-news-topic}} from the financial domain.

For each task, we sample 8 shots per class from the training set as the labeled data for SetFit and treat the whole original training set as the unlabeled data for DAPT. We run SetFit five times with different random seeds, which correspond to five different sets of few-shot samples. We report the average accuracy on the test set of each dataset over the five runs.

\section{Results}
\subsection{Training Order and DAPT Objectives}
\label{sec:objectives}
\begin{figure}[t]
    \centering
    \includegraphics[width=\columnwidth]{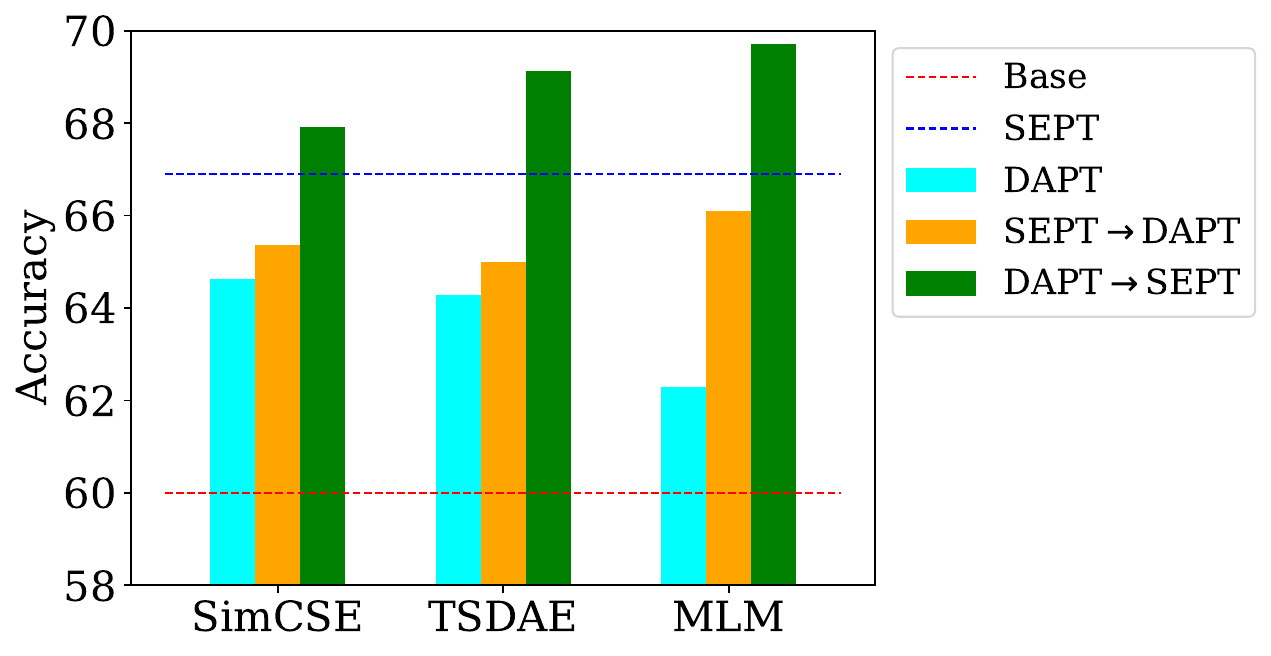}
    \caption{Averaged accuracy on 17 datasets of different DAPT training objectives (SimCSE, TSDAE, MLM) and different training strategies (without, before or after SEPT). Results on individual datasets are in \autoref{tbl:obj}.}
    \label{fig:objectives}
\end{figure}

In our first experiment, we compare two training orders: SEPT$\rightarrow$DAPT and DAPT$\rightarrow$SEPT, and three DAPT objectives: MLM, TSDAE and SimCSE. The results are shown in~\autoref{fig:objectives}.

Regarding the training order, DAPT$\rightarrow$SEPT outperforms SEPT$\rightarrow$DAPT for all three DAPT objectives. DAPT can enhance the SEPT baseline only when it is performed prior to SEPT, but this setting has the efficiency issue described in \autoref{sec:3.3}. On the other hand, DAPT has a negative impact on an already pre-trained sentence encoder, because it may distort the sentence representation space shaped by SEPT. These findings on our classification tasks are consistent with those on the retrieval tasks in~\citet{wang-etal-2021-tsdae-using}. 

With the DAPT$\rightarrow$SEPT order, MLM achieves the best result among three DAPT objectives, improving the SEPT baseline by around 3 points on average. Although TSDAE has been shown to have a clear advantage in tasks like re-ranking and paraphrase identification \citep{wang-etal-2021-tsdae-using}, it turns out to be suboptimal for sentence classification. 
On the contrary, MLM performs worse than TSDAE and SimCSE when there is no SEPT. We suppose that sentence classification with SetFit requires a good representation of both token- and sentence-level semantics, which are learned through MLM and SEPT respectively in the DAPT\textsubscript{MLM}$\rightarrow$SEPT setting. In other settings, either supervised sentence embedding training is absent (only DAPT), or token representation learning is missing (both TSDAE and SimCSE are for sentence representation learning).

\begin{table*}[t]
\small
\centering
\resizebox{\textwidth}{!}{
\begin{tabular}{llcccccccccccc} 
\hline
\begin{tabular}[c]{@{}l@{}}\textbf{Row }\\\textbf{No.}\end{tabular} & \textbf{Model} & \textbf{AC} & \textbf{BANK} & \textbf{AMI} & \textbf{AMS} & \textbf{MI} & \textbf{MD} & \textbf{EMO} & \textbf{IMDB} & \textbf{TSE} & \textbf{TC} & \textbf{ARM} & \textbf{Avg.} \\ 
\hline
\multicolumn{14}{c}{\textit{No SEPT}} \\
R1 & Base & 65.9 & 75.7 & 62.0 & 71.0 & 72.8 & 89.4 & 40.2 & 67.7 & 50.9 & 55.2 & 37.3 & 62.6 \\
R2 & DAPT & 69.4 & 80.4 & 70.0 & 79.4 & 80.6 & 94.7 & 37.3 & 74.9 & 55.1 & 48.2 & 41.9 & 66.5 \\ 
\hline
\multicolumn{14}{c}{\textit{Full SEPT}} \\
R3 & SEPT (prev. SOTA) & 76.1 & 77.0 & 66.8 & 73.3 & 78.4 & 90.6 & 52.2 & 84.8 & 63.2 & 63.6 & 44.2 & 70.0 \\
R4 & DAPT$\rightarrow$SEPT & 73.8 & \textbf{80.9} & 73.5 & 79.2 & \textbf{83.7} & 94.5 & 54.0 & 86.2 & 63.7 & 63.6 & 46.9 & 72.7 \\ 
\hline
\multicolumn{14}{c}{\textit{SEPT on Adapter}} \\
R5 & SEPT\textsubscript{ada} & 76.0 & 76.9 & 66.0 & 73.6 & 79.4 & 91.4 & 55.3 & 84.6 & 63.1 & 65.4 & 43.8 & 70.5 \\
R6 & DAPT$\rightarrow$SEPT\textsubscript{ada} & \textbf{77.9} & 80.7 & \textbf{73.8} & 79.3 & 82.9 & 94.7 & \textbf{54.7} & 85.6 & 65.0 & \textbf{65.5} & 47.0 & \textbf{73.4} \\
R7 & AdaSent & \textbf{77.9} & 80.6$^\dagger$ & 73.7$^\dagger$ & \textbf{80.5}$^\dagger$ & 82.7$^\dagger$ & \textbf{95.4}$^\dagger$ & 54.1 & \textbf{86.7} & \textbf{65.2} & 63.0 & \textbf{48.1}$^\dagger$ & \textbf{73.4} \\ 
\hline
\multicolumn{14}{c}{\textit{DAPT on Adapter}} \\
R8 & SEPT$\rightarrow$DAPT\textsubscript{ada} & 72.8 & 79.5 & 69.8 & 78.0 & 81.0 & 93.7 & 48.3 & 84.4 & 59.2 & 59.4 & 44.9 & 70.1 \\
\hline
\end{tabular}}
\caption{Classification accuracy on the MTEB classification tasks. Full SEPT means tuning all the PLM parameters in the sentence embedding pre-training. Best results on each dataset are in \textbf{bold}. $\dagger$ marks the cases where AdaSent outperforms SEPT (R5) with a statistical significance level of 0.05.
}
\label{tbl:main_table}
\end{table*}

\begin{table}[t]
\Huge
\centering
\resizebox{\columnwidth}{!}{
\begin{tabular}{llccccccc} 
\hline
\begin{tabular}[c]{@{}l@{}}\textbf{Row}\\\textbf{No.}\end{tabular} & \textbf{Model} & \textbf{FPB} & \textbf{TFNS} & \textbf{TFNT} & \textbf{ADE} & \textbf{RCT} & \textbf{LED} & \textbf{Avg.} \\ 
\hline
\multicolumn{9}{c}{\textit{No SEPT}} \\
R1 & Base & 49.2 & 51.1 & 57.7 & 60.7 & 49.6 & 64.2 & 55.4 \\
R2 & DAPT & 50.3 & 56.3 & 64.8 & 67.8 & 57.8 & 66.7 & 60.6 \\ 
\hline
\multicolumn{9}{c}{\textit{Full SEPT}} \\
R3 & SEPT & 63.0 & 65.0 & 62.2 & 62.3 & 61.5 & 65.6 & 63.3 \\
R4 & DAPT$\rightarrow$SEPT & 65.6 & 69.4 & 68.4 & 67.4 & 66.5 & \textbf{68.1} & 67.6 \\ 
\hline
\multicolumn{9}{c}{\textit{SEPT on Adapter}} \\
R5 & SEPT\textsubscript{ada} & 64.2 & 66.1 & 61.4 & 62.8 & 58.7 & 65.9 & 63.2 \\
R6 & DAPT$\rightarrow$SEPT\textsubscript{ada} & 66.1 & \textbf{69.9} & 68.5 & 65.8 & 67.4 & 68.0 & 67.6 \\
R7 & AdaSent & \textbf{66.4} & 69.8 & \textbf{68.6}$^\dagger$ & \textbf{67.8} & \textbf{67.5} & 67.8$^\dagger$ & \textbf{68.0} \\ 
\hline
\multicolumn{9}{c}{\textit{DAPT on Adapter}} \\
R8 & SEPT$\rightarrow$DAPT\textsubscript{ada} & 62.7 & 62.0 & 65.2 & 66.7 & 64.0 & 66.8 & 64.6 \\
\hline
\end{tabular}}
\caption{Classification accuracy on the domain-specific datasets. Best results on each dataset are in \textbf{bold}. $\dagger$ marks the cases where AdaSent outperforms SEPT (R5) with a statistical significance level of 0.05.} 
\label{tbl:main_table2}
\end{table}

\subsection{Combination of DAPT and SEPT}
\label{sec:dapt_and_sept}
In this subsection, we present the results of our main experiments on various combination strategies for DAPT and SEPT. The results on the MTEB datasets are reported in  \autoref{tbl:main_table}, and those for the domain-specific datasets are in \autoref{tbl:main_table2}. AdaSent achieves the best result on 10 out of 17 tasks, outperforming the not domain-adapted SEPT model by 3.9 on average on the MTEB datasets, and more prominently, by 4.7 on the datasets in \autoref{tbl:main_table2} with a larger domain shift from the pre-training data. The improvement is statistically significant on 8 datasets, with a significance level of 0.05. Our following analysis will focus on \autoref{tbl:main_table}, while similar trends can be observed in \autoref{tbl:main_table2}. 

SEPT is crucial to the final accuracy of classification methods based on sentence embeddings like SetFit, though this is not explicitly mentioned in the original SetFit paper~\cite{tunstall2022efficient}. SEPT improves both the Base model (R3 vs. R1) and the DAPT model (R4 vs. R2) by 7.3 and 6.2 points on average, respectively. 

By adding a DAPT stage before SEPT, the classification accuracy can be significantly increased by up to 6.7 points (on AMI) and 2.7 points on average (R4 vs. R3). 
However, as we discussed in~\autoref{sec:3.3}, executing the same SEPT procedure on every DAPT model results in computational inefficiency. 
As a more efficient alternative, our AdaSent avoids repeating SEPT by sharing a SEPT adapter across different downstream tasks, while obtaining comparable results without statistically significant difference (R7 vs. R4), except for the AMS dataset, where Adasent is even significantly better than DAPT→SEPT. 
The comparable performance of {DAPT$\rightarrow$SEPT}\textsubscript{ada} and AdaSent (R6 vs. R7) proves the viability of decoupling DAPT and SEPT: The SEPT adapter does not have to be trained on a specific DAPT model. 
Instead of doing SEPT on adapter, we also tried with DAPT on adapter (SEPT$\rightarrow$DAPT\textsubscript{ada}), which should be the most efficient method as explained in \autoref{sec:models}. Disappointingly, it can barely improve over SEPT (R8 vs. R3) and is much worse than AdaSent (R8 vs. R7). The reason could be that this setting suffers from the same problem as SEPT$\rightarrow$DAPT, as the DAPT phase, despite on an adapter, is still conducted after SEPT.  

\subsection{Comparison of PEFT Methods}
\label{sec:peft}
\begin{figure}[t]
    \centering
    \includegraphics[width=\columnwidth]{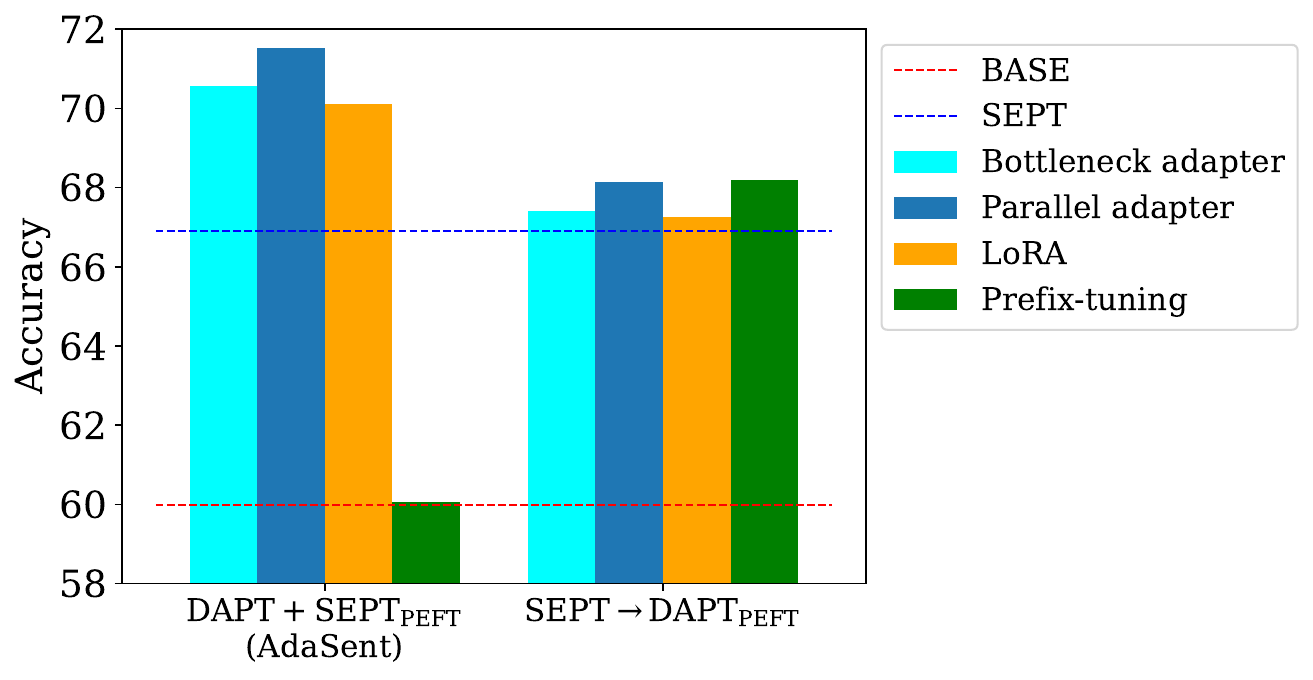}
    \caption{Averaged accuracy of different PEFT methods. SEPT\textsubscript{PEFT} stands for SEPT on a PEFT module. More detailed results are available in \autoref{tbl:peft}.} 
    \label{fig:peft}
\end{figure}
We experimented with four different PEFT methods for both SEPT and DAPT (\autoref{fig:peft}). When applied to SEPT in AdaSent, parallel adapter works best on the majority of the datasets (\autoref{tbl:peft}) and on average. Prefix-tuning is significantly worse than the other three methods. This might be due to the fact that the data in our SEPT dataset NLI+SC+SE come from three different tasks, whose properties cannot be compressed into a single prefix. 
When applied to DAPT in the  SEPT$\rightarrow$DAPT\textsubscript{PEFT} setting, their performance exhibits variability across different datasets (\autoref{tbl:peft}), but none of the four PEFT methods in this setting can beat the AdaSent variants due to the critical drawback of the setting as discussed at the end of \autoref{sec:dapt_and_sept}. 

\subsection{Tunable Parameters in SetFit}
\begin{table}[t]
\small
\centering
\resizebox{\columnwidth}{!}{
\begin{tabular}{lcccc} 
\hline
\begin{tabular}[c]{@{}l@{}}\textbf{Tunable }\\\textbf{Parameters}\end{tabular} & \begin{tabular}[c]{@{}c@{}}None\\(0\%)\end{tabular} & \begin{tabular}[c]{@{}c@{}}Adapter\\(4\%)\end{tabular} & \begin{tabular}[c]{@{}c@{}}Transformer\\(96\%)\end{tabular} & \begin{tabular}[c]{@{}c@{}}All\\(100\%)\end{tabular} \\ 
\hline
\textbf{Accuracy} & 65.0 & 69.0 & 71.2 & 71.5 \\
\hline
\end{tabular}}
\caption{Results of tuning subsets of model parameters (marked with relative sizes) in the final SetFit stage of AdaSent. None means only training the logistic regression head.} 
\label{tbl:tune_part}
\end{table}

\begin{table}[t]
\small
\centering
\begin{tabular}{lll} 
\hline
 & \textbf{MLM} & \textbf{TSDAE}\\ 
\hline
{DAPT+SEPT}\textsubscript{ada} (AdaSent) & 69.7 & 67.3 \\
{DAPT$\rightarrow$SEPT} & 69.7 & 69.1 \\
\hline
\end{tabular}
\caption{Averaged accuracy of different DAPT objectives in AdaSent and DAPT$\rightarrow$SEPT.} 
\label{tbl:tsdae_mlm}
\end{table}

\begin{table}[t]
\small
\centering
\begin{tabular}{lll} 
\hline
\textbf{Self-training} & \textbf{No} & \textbf{Yes} \\ 
\hline
SEPT & 67.6 & 68.6 (+1.0) \\
DAPT+SEPT\textsubscript{ada} (AdaSent) & 71.5 & 72.4 (+0.9) \\
\hline
\end{tabular}
\caption{Averaged accuracy of AdaSent and SEPT, w/ or w/o self-training.} 
\label{table:self_train}
\end{table}

We tune various subsets of parameters in the SetFit stage of AdaSent and compare the results in~\autoref{tbl:tune_part}. We found that only updating the adapter parameters is not sufficient. However, tuning only the Transformer backbone leads to almost the same results as tuning all parameters (i.e. Transformer + adapter). 
This indicates that with only few-shot labeled data, SetFit must at least update the Transformer parameters to achieve good performance, and cannot work well on an adapter as in the case of SEPT, where much more supervised data are available. 

\subsection{Explaining the Success of AdaSent}
\label{sec:explain_reason}
The success of AdaSent relies on the fact that a SEPT adapter trained on a base PLM can be unproblematically inserted into any domain-adapted version of the same PLM. This might be because in both original pre-training and domain-adaptive pre-training, the PLM parameters are consistently tuned with the MLM objective. This implies that the adapter can generalize to work together with PLM parameters trained on different types of data, from general-language data (e.g. BookCorpus,~\citealp{book_corpus}) to domain-specific data, as long as the same MLM objective is used. 
To verify this idea, we replace the MLM objective with TSDAE in both AdaSent and DAPT$\rightarrow$SEPT. As shown in~\autoref{tbl:tsdae_mlm}, using TSDAE instead of MLM in the DAPT stage of AdaSent leads to a substantial decrease of 2.4 points in the classification accuracy, while the performance drop in DAPT$\rightarrow$SEPT is relatively marginal (0.6 on average). This supports our hypothesis that the adapter can only generalize to collaborate with PLM parameters that are domain-adapted with the same objective as in the pre-training.

\begin{table}[t]
\centering
\resizebox{\columnwidth}{!}{
\begin{tabular}{lccccc} 
\hline
\multirow{2}{*}{\textbf{Method}} & \multirow{2}{*}{\begin{tabular}[c]{@{}c@{}}\textbf{DAPT}\\\textbf{Steps}\end{tabular}} & \multicolumn{3}{c}{\textbf{Cost (hour)}} & \multirow{2}{*}{\textbf{Acc.}} \\ 
\cline{3-5}
 &  & \textbf{SEPT} & \textbf{DAPT} & \textbf{Total} &  \\ 
\hline
\multirow{5}{*}{DAPT$\rightarrow$SEPT} & 0 & \multirow{5}{*}{15 $\times$ 0.27} & 0.00 & 4.05 & 67.6 \\
 & 100 &  & 0.44 & 4.49 & 69.8 \\
 & 500 &  & 2.21 & 6.26 & 70.7 \\
 & 1000 &  & 4.42 & 8.47 & \multicolumn{1}{l}{71.1} \\
 & 2000 &  & 8.83 & 12.88 & 71.5 \\ 
\hline
\multirow{5}{*}{AdaSent} & 0 & \multirow{5}{*}{1 $\times$ 0.17} & 0.00 & 0.17 & 67.9 \\
 & 100 &  & 0.44 & 0.61 & 69.5 \\
 & 500 &  & 2.21 & 2.38 & 70.7 \\
 & 1000 &  & 4.42 & 4.59 & \multicolumn{1}{l}{71.2} \\
 & 2000 &  & 8.83 & 9.00 & 71.4 \\
\hline
\end{tabular}}
\caption{Total training cost on 15 tasks with {DistilRoBERTa} as base PLM on a Tesla V100 GPU.}
\label{tbl:cost}
\end{table}

\subsection{Combining DAPT and Self-Training}
Besides DAPT, another major way to utilize the unlabeled data is self-training, which has been shown to be complementary to DAPT~\cite{li-etal-2021-task-adaptive}. To integrate self-training into SetFit, we first encode the unlabeled data with the sentence encoder (in our case a DAPT Transformer + SEPT adapter) trained with few-shot labeled data in the contrastive fine-tuning phase. When training the classification head, we iteratively pseudo-label the encoded unlabeled sentences and train with both the pseudo-labeled and the gold-labeled data\footnote{The training details are available in \autoref{sec:self_train_setting}.}.
In \autoref{table:self_train}, we show that self-training can further improve both SEPT and AdaSent's accuracy by 1.0 and 0.9 on average, respectively. These two close improvements reveal that the benefit of self-training is orthogonal to that of AdaSent/DAPT. We leave more complex approaches of combining AdaSent and self-training for future work.

\section{Training Cost}
\autoref{tbl:cost} gives an overview of the training cost for {DAPT$\rightarrow$SEPT} and AdaSent in our experiments. We use a Tesla V100 GPU for training. We leave out IMDB and LED as they have too long sequences (cf.~\autoref{tbl:dataset}), thus cannot represent the majority of our tasks.

With AdaSent, SEPT is trained once for 0.17h and the SEPT adapter can be shared across tasks. In contrast, DAPT$\rightarrow$SEPT costs 0.27 hours additionally for every task due to its repeated SEPT. In our experiments, we use relatively small-sized data for SEPT. However, the SEPT cost can increase dramatically if much larger training data are used. For example, SEPT on the combination of all datasets in \autoref{tbl:sept_data} for 1 epoch can take 4 hours, resulting in 15 $\times$ 4 hours for DAPT$\rightarrow$SEPT for 15 tasks. For DAPT, we can see that 1000 steps are already sufficient for a substantial improvement in accuracy. In this case, AdaSent takes 4.59 hours for the training on 15 tasks in total, while DAPT$\rightarrow$SEPT takes 8.47 hours ($\times$1.85).

\section{Conclusion}
We introduce an efficient method to obtain domain-adapted sentence embeddings for few-shot classification. We found that SetFit, the previous state-of-the-art approach, can be significantly improved by introducing a simple Domain-Adaptive Pre-Training (DAPT) stage before its Sentence-Embedding Pre-Training (SEPT). However, this DAPT$\rightarrow$SEPT approach requires the same SEPT procedure to be done on each DAPT-ed PLM for every domain, resulting in computational inefficiency. We propose a novel approach, AdaSent, to address this issue by storing the SEPT knowledge in an adapter that is trained on an unadapted PLM and insertable into any DAPT-ed PLM. AdaSent matches or surpasses the performance of DAPT$\rightarrow$SEPT, while significantly reducing the training cost of SEPT. We attribute the success of AdaSent to the generalization ability of the SEPT adapter to work with PLM parameters trained on data from different domains with a consistent MLM objective.

\section*{Limitations}
Since our method is based on SetFit, it inherits some of its limitations. It is, for example, not applicable for sentence pair classification like NLI. In addition, the advantage of SetFit is not significant in classification tasks with too many classes. Moreover, as our method is based on sentence embeddings, its application is limited to sentence classification, unlike other few-shot classification methods that can also handle token-level classification tasks like NER and POS tagging. 

Another limitation is associated with the fact that the SEPT adapter in our method can only be inserted into domain-adapted language models with the same unmodified tokenizer and vocabulary as the original base PLM. For DAPT-ed models with a domain-specific tokenizer or vocabulary, we suppose the adapter trained on the original PLM will not be compatible anymore. 

\section*{Ethics Statement}
Our experiments use publicly available datasets and benchmarks for training and evaluation, which are commonly used in the field of NLP. No personal information or sensitive data are involved in our work. Existing biases in the datasets or pre-trained models can still be relevant concerns, since we do not specifically focus on mitigating them in the current work. 

\section*{Acknowledgements}
This work has been funded by HUAWEI Technologies (Ireland) Co., Ltd. and by the German Federal Ministry of Education and Research (BMBF) under the promotional reference 13N15897 (MISRIK).

\bibliography{anthology,custom}
\bibliographystyle{acl_natbib}

\appendix

\section{Implementation}
See \autoref{tbl:code} for the implementation of methods used in this work. 

\label{sec:code}

\section{Experiment with SEPT Datasets}
\label{sec:sept_data}
We experiment with different SEPT datasets\footnote{See \url{https://www.sbert.net/examples/training/paraphrases/README.html} for information of the datasets.} to check their transferability to downstream tasks \autoref{tbl:sept_data}. On average, AllNLI, SentenceCompression and StackexchangeDuplicateQuestions are the top three datasets. The similarity between the SEPT data and the downstream data seems to have an influence on the performance. For example, QA-related data (YahooAnswersTitleAnswer, StackexchangeDuplicateQuestions and  YahooAnswersQuestionAnswer) are especially beneficial for the classification tasks involving user utterances in dialogues (BANK, AMI, AMS, MI, MD). Given this observation, one might want to search for the optimal SEPT datasets depending on certain types of classification tasks. Our adapter-based method enables efficient SEPT, which helps to ease the data selection. 

\section{DAPT Objectives and Training Order}
Results on individual datasets are listed in \autoref{tbl:obj}.

\section{Results on DistilBERT}
\label{sec:distilbert}
We report the results on DistilBERT in \autoref{tbl:distilbert}. Similar to DistilRoBERTa, DAPT with MLM (DAPT$\rightarrow$SEPT and DAPT+SEPT\textsubscript{ada}) improves the performance of SEPT by around 3 points on average. Replacing full SEPT with SEPT adapter causes a slight drop of around 0.5 in the classification accuracy. Interestingly, without any supervised sentence embedding pre-training, DAPT itself can outperform SEPT on some datasets (AC, ADE, LED).

\begin{table}[t]
\small
\centering
\begin{tabularx}{\columnwidth}{lX} 
\hline
\textbf{Method} & \textbf{Used Implementation} \\ 
\hline
PEFT & \url{https://github.com/adapter-hub/adapter-transformers} \\
TSDAE & \url{https://github.com/UKPLab/sentence-transformers} \\
SEPT & \url{https://github.com/UKPLab/sentence-transformers} \\
SimCSE & \url{https://github.com/princeton-nlp/SimCSE} \\
MLM & \url{https://github.com/huggingface/transformers/blob/main/examples/pytorch/language-modeling/run_mlm_no_trainer.py} \\
SetFit & \url{https://github.com/huggingface/setfit} \\
\hline
\end{tabularx}
\caption{Implementation used in this work.}
\label{tbl:code}
\end{table}

\begin{table*}[t]
\centering
\resizebox{\textwidth}{!}{
\begin{tabular}{lllllllllllll}
\hline
\textbf{SEPT Data} & \textbf{AC} & \textbf{BANK} & \textbf{AMI} & \textbf{AMS} & \textbf{MI} & \textbf{MD} & \textbf{EMO} & \textbf{IMDB} & \textbf{TSE} & \textbf{TC} & \textbf{ARM} & \textbf{Avg.} \\
\hline
AllNLI & 65.5 & \textbf{75.5} & \uline{63.5} & 72.1 & 74.2 & 90.2 & \uline{48.8} & \textbf{84.9} & \textbf{62.7} & \uline{64.8} & \textbf{43.5} & \uline{67.8} \\
SentenceCompression & \textbf{74.0} & 74.9 & 61.5 & 72.9 & 74.9 & 90.6 & \textbf{52.4} & \uline{83.8} & \uline{60.3} & 58.7 & 42.2 & \textbf{67.9} \\
SimpleWiki & 65.4 & 74.3 & 59.4 & 71.1 & 70.8 & 89.3 & 45.3 & 83.5 & 59.7 & 63.8 & 41.9 & 65.9 \\
Altlex & 68.4 & 74.5 & 59.6 & 71.1 & 72.0 & 89.2 & 45.6 & 81.8 & 57.7 & 62.6 & 41.6 & 65.8 \\
QuoraDuplicatesTriplets & \uline{73.6} & \uline{75.3} & 60.9 & 71.1 & 75.0 & 89.5 & 44.6 & 81.0 & 58.0 & 61.8 & 42.0 & 66.6 \\
CocoCaptions & 58.4 & 74.3 & 58.7 & 71.2 & 71.6 & 89.6 & 45.5 & 60.3 & 51.2 & 58.4 & 37.7 & 61.5 \\
Flickr30kCaptions & 58.0 & 74.1 & 59.5 & 71.2 & 73.8 & 89.4 & 45.1 & 60.0 & 52.3 & \textbf{66.4} & 36.8 & 62.4 \\
YahooAnswersTitleQuestion & 69.0 & 75.2 & 60.5 & 72.4 & 75.5 & 90.6 & 46.3 & 83.4 & 55.1 & 58.6 & 41.6 & 66.2 \\
YahooAnswersTitleAnswer & 71.5 & 75.1 & 61.2 & \textbf{73.7} & \uline{75.5} & \textbf{90.9} & 44.9 & 80.9 & 55.6 & 52.1 & 41.5 & 65.7 \\
StackexchangeDuplicateQuestions & 72.0 & 75.1 & \textbf{64.0} & \uline{73.6} & \textbf{77.9} & 90.2 & 46.1 & 77.2 & 58.2 & 60.6 & \uline{43.1} & 67.1 \\
YahooAnswersQuestionAnswer & 67.3 & 75.0 & 61.2 & 73.4 & 75.4 & \uline{90.8} & 45.3 & 81.5 & 52.3 & 62.7 & 41.2 & 66.0 \\
\hline
\end{tabular}}
\caption{Results on MTEB tasks of {SEPT} model trained on different datasets. The best scores are marked in bold and second best with underline. We sample 100K instances from each SEPT dataset and train for 1 epoch. }
\label{tbl:sept_data}
\end{table*}

\begin{table*}[t]
\centering
\resizebox{\textwidth}{!}{
\begin{tabular}{lllllllllllllllllll} 
\hline
\textbf{Model} & \textbf{AC} & \textbf{BANK} & \textbf{AMI} & \textbf{AMS} & \textbf{MI} & \textbf{MD} & \textbf{EMO} & \textbf{IMDB} & \textbf{TSE} & \textbf{TC} & \textbf{ARM} & \textbf{FPB} & \textbf{TFNS} & \textbf{TFNT} & \textbf{ADE} & \textbf{RCT} & \textbf{LED} & \textbf{Avg.} \\ 
\hline
\multicolumn{19}{c}{\textit{TSDAE }} \\
{DAPT} & 74.1 & \uline{77.6} & 64.4 & 76.0 & 77.4 & 93.0 & 46.3 & 78.4 & 53.7 & 49.6 & 44.8 & 51.1 & 55.7 & 61.6 & 62.5 & 58.8 & 67.8 & 64.3 \\
{SEPT$\rightarrow$DAPT} & \textbf{79.5} & 77.5 & 66.8 & 76.4 & \uline{79.9} & 93.2 & 46.1 & 79.5 & 50.6 & 47.6 & 47.3 & 58.4 & 53.2 & 63.0 & 60.0 & 57.6 & \textbf{68.3} & 65.0 \\
{DAPT$\rightarrow$SEPT} & 72.9 & 77.3 & \textbf{67.9} & \uline{76.3} & 79.8 & \uline{93.4} & \textbf{52.0} & \textbf{85.6} & 63.6 & 62.1 & \uline{48.9} & \uline{66.6} & \textbf{63.9} & \uline{66.2} & \uline{65.3} & 65.4 & \uline{68.1} & \uline{69.1} \\ 
\hline
\multicolumn{19}{c}{\textit{SimCSE }} \\
{DAPT} & 71.2 & 75.3 & 61.9 & 73.0 & 75.6 & 89.9 & 46.8 & 78.5 & 59.2 & 57.2 & 44.5 & 58.8 & 57.3 & 60.5 & 61.3 & 63.6 & 64.3 & 64.6 \\
{SEPT$\rightarrow$DAPT} & 66.2 & 75.4 & 63.3 & 74.0 & 78.8 & 89.8 & 44.4 & 66.9 & 59.1 & \textbf{69.9} & 44.7 & 64.9 & 61.7 & 60.2 & 58.3 & \textbf{68.2} & 65.5 & 65.4 \\
{DAPT$\rightarrow$SEPT} & 71.4 & 76.0 & 65.4 & 73.9 & 78.8 & 91.9 & 48.4 & 84.7 & \textbf{64.1} & 66.4 & 46.9 & \textbf{66.9} & 62.5 & 62.3 & 62.7 & 66.1 & 66.0 & 67.9 \\ 
\hline
\multicolumn{19}{c}{\textit{MLM }} \\
{DAPT} & 61.8 & 76.6 & 61.8 & 75.7 & 77.0 & 93.2 & 38.0 & 64.8 & 52.7 & 60.3 & 44.2 & 53.2 & 58.3 & 63.9 & 63.9 & 46.4 & 67.3 & 62.3 \\
{SEPT$\rightarrow$DAPT} & 73.2 & 76.4 & 65.0 & 75.6 & 79.0 & 92.8 & 49.1 & 79.3 & 58.3 & 51.3 & 48.8 & 55.9 & 61.5 & 65.9 & 64.7 & 59.2 & 67.9 & 66.1 \\
{DAPT$\rightarrow$SEPT} & 72.7 & \textbf{78.0} & \uline{67.4} & \textbf{77.0} & \textbf{82.4} & \textbf{93.7} & 49.9 & \uline{85.5} & \uline{63.9} & \uline{65.3} & \textbf{50.9} & \uline{66.6} & \uline{63.8} & \textbf{66.8} & \textbf{66.5} & \uline{66.9} & 67.7 & \textbf{69.7} \\ 
\hline
\multicolumn{19}{c}{\textit{Baselines }} \\
{SEPT} & 70.2 & 75.5 & 64.5 & 73.6 & 77.4 & 90.6 & \uline{51.8} & 84.2 & 63.3 & 61.8 & 43.3 & 65.5 & 60.8 & 61.8 & 64.0 & 64.3 & 64.6 & 66.9 \\
{Base} & 65.9 & 75.2 & 60.6 & 71.0 & 73.9 & 89.4 & 40.3 & 68.0 & 50.9 & 55.6 & 37.6 & 48.5 & 50.8 & 57.8 & 60.9 & 49.2 & 64.2 & 60.0 \\
\hline
\end{tabular}}
\caption{Comparison of different DAPT objectives and training order of DAPT and SEFT. The best scores are marked in bold and second best with underline. Note that the training settings of DAPT here is different from that in~\autoref{tbl:main_table} and~\autoref{tbl:main_table2}: We do DAPT for 3 epochs instead of a fixed-number of steps.}
\label{tbl:obj}
\end{table*}

\begin{table*}[t]
\centering
\resizebox{\textwidth}{!}{
\begin{tabular}{lrrrrrrrrrrrrrrrrrr}
\hline
\textbf{Model} & \multicolumn{1}{l}{\textbf{AC}} & \multicolumn{1}{l}{\textbf{BANK}} & \multicolumn{1}{l}{\textbf{AMI}} & \multicolumn{1}{l}{\textbf{AMS}} & \multicolumn{1}{l}{\textbf{MI}} & \multicolumn{1}{l}{\textbf{MD}} & \multicolumn{1}{l}{\textbf{EMO}} & \multicolumn{1}{l}{\textbf{IMDB}} & \multicolumn{1}{l}{\textbf{TSE}} & \multicolumn{1}{l}{\textbf{TC}} & \multicolumn{1}{l}{\textbf{ARM}} & \multicolumn{1}{l}{\textbf{FPB}} & \multicolumn{1}{l}{\textbf{TFNS}} & \multicolumn{1}{l}{\textbf{TFNT}} & \multicolumn{1}{l}{\textbf{ADE}} & \multicolumn{1}{l}{\textbf{RCT}} & \multicolumn{1}{l}{\textbf{LED}} & \multicolumn{1}{l}{\textbf{Avg.}} \\
\hline
\multicolumn{19}{c}{\textit{No SEPT}} \\
{Base} & 74.1 & 73.4 & 61.9 & 70.6 & 75.6 & 88.2 & 35.3 & 64.5 & 49.2 & 58.0 & 37.9 & 49.6 & 43.4 & 52.6 & 64.7 & 53.2 & 64.3 & 59.8 \\
{DAPT} & \textbf{82.1} & 79.6 & 70.2 & 79.2 & 82.9 & 95.1 & 40.1 & 78.3 & 56.1 & 50.6 & 45.2 & 53.8 & 53.2 & 65.6 & \textbf{72.4} & 65.3 & \textbf{67.3} & 66.9 \\
\hline
\multicolumn{19}{c}{\textit{Full SEPT}} \\
{SEPT} & 72.6 & 75.7 & 67.5 & 74.3 & 79.7 & 91.9 & 48.4 & 80.7 & 63.8 & 63.9 & 42.8 & 61.1 & 62.1 & 58.2 & 63.9 & 58.6 & 65.6 & 66.5 \\
{DAPT$\rightarrow$SEPT} & 75.6 & 79.7 & \textbf{73.9} & \textbf{80.4} & \textbf{83.9} & 94.9 & 50.9 & \textbf{83.4} & \textbf{63.8} & \textbf{64.4} & 46.1 & \textbf{62.2} & 64.6 & \textbf{66.3} & 65.9 & 64.0 & \textbf{67.3} & \textbf{69.8} \\
\hline
\multicolumn{19}{c}{\textit{SEPT on Adapter}} \\
{SEPT}\textsubscript{ada} & 75.5 & 75.2 & 66.3 & 73.7 & 78.5 & 91.4 & 50.0 & 78.9 & 63.5 & 58.2 & 42.1 & 63.4 & 61.8 & 55.9 & 62.3 & 58.2 & 65.3 & 65.9 \\
AdaSent & 80.6 & \textbf{79.8} & 72.1 & 80.3 & 82.4 & \textbf{95.7} & \textbf{51.6} & 82.2 & 62.8 & 56.6 & \textbf{47.1} & 60.4 & \textbf{66.2} & 63.4 & 64.9 & \textbf{65.8} & \textbf{67.3} & 69.4 \\
\hline
\end{tabular}}
\caption{Results on DistilBERT. Best scores are in bold.}
\label{tbl:distilbert}
\end{table*}

\section{PEFT results}
\label{sec:peft_appendix}
Results on individual datasets when using different PEFT methods as discussed in \autoref{sec:peft} in our AdaSent method ({DAPT+SEPT}\textsubscript{PEFT}) and {SEPT$\rightarrow$DAPT}\textsubscript{PEFT} are shown in
\autoref{tbl:peft}. 

\begin{table*}[t]
\centering
\resizebox{\textwidth}{!}{
\begin{tabular}{lllllllllllllllllll} 
\hline
\textbf{PEFT method} & \textbf{AC} & \textbf{BANK} & \textbf{AMI} & \textbf{AMS} & \textbf{MI} & \textbf{MD} & \textbf{EMO} & \textbf{IMDB} & \textbf{TSE} & \textbf{TC} & \textbf{ARM} & \textbf{FPB} & \textbf{TFNS} & \textbf{TFNT} & \textbf{ADE} & \textbf{RCT} & \textbf{LED} & \textbf{Avg.} \\ 
\hline
\multicolumn{19}{c}{{DAPT+SEPT}\textsubscript{PEFT} (AdaSent)} \\
Bottleneck adapter & 76.3 & \textbf{80.7} & 73.0 & 80.2 & 82.2 & 95.3 & 51.5 & \textbf{87.4} & 64.2 & 58.6 & 48.0 & 65.0 & 66.7 & \textbf{68.6} & 65.2 & \textbf{68.7} & \textbf{68.2} & 70.6 \\
Parallel adapter & 77.9 & 80.6 & \textbf{73.7} & \textbf{80.5} & \textbf{82.7} & 95.4 & \textbf{54.1} & 86.7 & \textbf{65.2} & \textbf{63.0} & \textbf{48.1} & \textbf{66.4} & \textbf{69.8} & \textbf{68.6} & \textbf{67.8} & 67.5 & 67.8 & \textbf{71.5} \\
LoRA & \textbf{79.0} & \textbf{80.7} & 72.9 & 79.6 & 82.0 & \textbf{94.6} & 52.9 & 85.1 & 63.1 & 60.1 & 46.3 & 63.5 & 64.7 & 66.9 & 67.1 & 65.7 & 67.5 & 70.1 \\
Prefix-tuning & 53.1 & 80.1 & 69.8 & 78.5 & 80.3 & 94.1 & 41.8 & 62.9 & 49.0 & 46.7 & 36.0 & 41.8 & 47.9 & 65.3 & 58.8 & 47.6 & 67.3 & 60.0 \\ 
\hline
\multicolumn{19}{c}{{SEPT$\rightarrow$DAPT}\textsubscript{PEFT}} \\
Bottleneck Adapter & 76.4 & 77.5 & 67.0 & 74.6 & 79.6 & 91.5 & 50.3 & 82.9 & 61.4 & 58.0 & 43.5 & 60.7 & 63.7 & 63.6 & 64.6 & \textbf{65.1} & 65.8 & 67.4 \\
Parallel Adapter & 72.8 & \textbf{79.5} & \textbf{69.8} & \textbf{78.0} & \textbf{81.0} & \textbf{93.7} & 48.3 & 84.4 & 59.2 & 59.4 & \textbf{44.9} & 62.7 & 62.0 & \textbf{65.2} & \textbf{66.7} & 64.0 & \textbf{66.8} & 68.1 \\
LoRA & 77.5 & 77.3 & 66.7 & 73.6 & 78.5 & 91.2 & 51.0 & 83.8 & \textbf{63.2} & 58.9 & 43.6 & 61.1 & 60.3 & 63.6 & 63.3 & 64.2 & 65.5 & 67.3 \\
Prefix-tuning & \textbf{78.9} & 77.3 & 66.0 & 72.6 & 78.6 & 90.5 & \textbf{52.9} & \textbf{84.7} & 62.6 & \textbf{63.7} & 44.2 & \textbf{64.5} & \textbf{67.7} & 62.7 & 63.8 & 63.0 & 65.8 & \textbf{68.2} \\
\hline
\end{tabular}}
\caption{Results on individual datasets of different PEFT methods for {DAPT+SEPT}\textsubscript{PEFT} (AdaSent) and {SEPT$\rightarrow$DAPT}\textsubscript{PEFT}. Best scores are in bold for both models.} 
\label{tbl:peft}
\end{table*}

\section{Evaluation Datasets}
\label{sec:dataset_appendix}
\autoref{tbl:data_example} provides examples from each evaluation dataset.  

\begin{table*}[t]
\small
\centering
\begin{tabularx}{\textwidth}{llXl} 
\hline
\textbf{Dataset} & \textbf{Abbr.} & \textbf{Text} & \textbf{Label} \\ 
\hline
\multicolumn{4}{c}{\textit{MTEB classification}} \\ 
\hline
\begin{tabular}[c]{@{}l@{}}Amazon\\Counterfactual \\ \citep{oneill-etal-2021-wish_counterfactual}\end{tabular} & AC & In person it looks as though it
  would have cost a lot more. & counterfactual \\ 
\hline
Banking77 \\ \citep{casanueva-etal-2020-efficient_banking77} & BANK & I am still waiting on my card? & card\_arrival \\ 
\hline
\begin{tabular}[c]{@{}l@{}}Amazon \\Massive Intent \\ \citep{fitzgerald-etal-2023-massive}\end{tabular} & AMI & wake me up at nine am on friday & alarm\_set \\ 
\hline
\begin{tabular}[c]{@{}l@{}}Amazon \\Massive Scenario \\ \citep{fitzgerald-etal-2023-massive}\end{tabular} & AMS & wake me up at nine am on friday & alarm \\ 
\hline
MTOP Intent \\ \citep{li-etal-2021-mtop}~ & MI & Has Angelika Kratzer video
  messaged me? & GET\_MESSAGE \\ 
\hline
MTOP Domain \\ \citep{li-etal-2021-mtop} & MD & Has Angelika Kratzer video
  messaged me? & messaging \\ 
\hline
Emotion \\ \citep{saravia-etal-2018-carer_emotion}& EMO & ive been feeling a little burdened lately wasnt sure why that was & sadness \\ 
\hline
Imdb \\ \citep{maas-etal-2011-learning_imdb} & IMDB & I rented I AM CURIOUS-YELLOW from my video store because of all the controversy that surrounded it when it was first released in 1967. I also heard that at first it was seized by U.S. customs if it ever tried to enter this country, therefore being a fan of films considered "controversial" I really had to see this for myself.\textless{}br /\textgreater{}\textless{}br /\textgreater{}The plot is centered around a young Swedish drama student named Lena who wants to learn everything she can about life. [...] I AM CURIOUS-YELLOW is a good film for anyone wanting to study the meat and potatoes (no pun intended) of Swedish cinema. But really, this film doesn't have much of a plot. & negative \\ 
\hline
\begin{tabular}[c]{@{}l@{}}Twitter \\Sentiment \\Extraction\end{tabular} & TSE & ~I\`{}d have responded, if I were going & neutral \\ 
\hline
\begin{tabular}[c]{@{}l@{}}Toxic \\Conversation\end{tabular} & TC & theres not enough going on
  around here for air service none want to waste there time on this town & not toxic \\ 
\hline
\begin{tabular}[c]{@{}l@{}}Amazon \\Reviews Multi \\ \citep{Amazon_reviews} \end{tabular} & ARM & I
  received my first order of this product and it was broke so I ordered it
  again. The second one was broke in more places than the first. I can't blame
  the shipping process as it's shrink wrapped and boxed. & 0 \\ 
\hline
\multicolumn{4}{c}{\textit{Domain-specific tasks}} \\ 
\hline
\begin{tabular}[c]{@{}l@{}}Financial \\PhraseBank\end{tabular} & FPB & With
  the new production plant the company would increase its capacity to meet the
  expected increase in demand and would improve the use of raw materials and
  therefore increase the production profitability . & Positive \\ 
\hline
\begin{tabular}[c]{@{}l@{}}Twitter \\Financial \\News \\Sentiment\end{tabular} & TFNS & Grubhub
  gains a bear on margin view & Bearish \\ 
\hline
\begin{tabular}[c]{@{}l@{}}Twitter \\Financial \\News Topic\end{tabular} & TFNT & Analysts
  reveal the top stocks with 'significant upside potential' heading into
  earnings https://t.co/lfaLK3nwAz & Analyst Update \\ 
\hline
\begin{tabular}[c]{@{}l@{}}Adverse \\Drug \\Events\end{tabular} & ADE & Intravenous
  azithromycin-induced ototoxicity. & Related \\ 
\hline
\begin{tabular}[c]{@{}l@{}}PubMed \\RCT\end{tabular} & RCT & Outcome
  measures included pain reduction and improvement in function scores and
  systemic inflammation markers . & Methods \\ 
\hline
LEDGAR & LED & Except
  as otherwise set forth in this Debenture, the Company, for itself and its
  legal representatives, successors and assigns, expressly waives presentment,
  protest, demand, notice of dishonor, notice of nonpayment, notice of
  maturity, notice of protest, presentment for the purpose of accelerating
  maturity, and diligence in collection. & Waivers \\
\hline
\end{tabularx}
\caption{Examples from evaluation datasets.}
\label{tbl:data_example}
\end{table*}

\section{Self-Training Setting}
\label{sec:self_train_setting}
In the SetFit phase, we contrastively fine-tune the sentence embedding model with the few-shot data as before (\autoref{sec:setfit}), but replace the normal Logistic Regression fitting with self-training on both labeled and unlabeled data. For this, we use the \texttt{SelfTrainingClassifier} from scikit-learn\footnote{\url{https://scikit-learn.org/stable/modules/generated/sklearn.semi_supervised.SelfTrainingClassifier.html}} with 10 iterations and a threshold of 0.9. At each iteration, the classifier predicts the label of the unlabeled data. The pseudo-labeled data with a confidence score higher than the threshold are used to augment the training data in the next iteration.    

\end{document}